# Contrastive Learning for Predicting Cancer Prognosis Using Gene Expression Values

Anchen Sun[1]   Elizabeth J. Franzmann[2,4]   Zhibin Chen[3,4]   Xiaodong Cai[1,4,5]

[1]Department of Electrical and Computer Engineering, University of Miami, Florida, USA
[2] Department of Otolaryngology, University of Miami, Florida, USA
[3]Department of Microbiology and Immunology, University of Miami, Florida, USA
[4]Sylvester Comprehensive Cancer Center, University of Miami, Florida, USA
[5]Correspondence: x.cai@miami.edu

**Abstract**

Recent advancements in image classification have demonstrated that contrastive learning (CL) can aid in further learning tasks by acquiring good feature representation from a limited number of data samples. In this paper, we applied CL to tumor transcriptomes and clinical data to learn feature representations in a low-dimensional space. We then utilized these learned features to train a classifier to categorize tumors into a high- or low-risk group of recurrence. Using data from The Cancer Genome Atlas (TCGA), we demonstrated that CL can significantly improve classification accuracy. Specifically, our CL-based classifiers achieved an area under the receiver operating characteristic curve (AUC) greater than 0.8 for 14 types of cancer, and an AUC greater than 0.9 for 2 types of cancer. We also developed CL-based Cox (CLCox) models for predicting cancer prognosis. Our CLCox models trained with the TCGA data outperformed existing methods significantly in predicting the prognosis of 19 types of cancer under consideration. The performance of CLCox models and CL-based classifiers trained with TCGA lung and prostate cancer data were validated using the data from two independent cohorts. We also show that the CLCox model trained with the whole transcriptome significantly outperforms the Cox model trained with the 21 genes of Oncotype DX that is in clinical use for breast cancer patients. CL-based classifiers and CLCox models for 19 types of cancer are publicly available and can be used to predict cancer prognosis using the RNA-seq transcriptome of an individual tumor. Python codes for model training and testing are also publicly accessible, and can be applied to train new CL-based models using gene expression data of tumors.

## 1   Introduction

Numerous gene prognostic signatures have been developed over the last two decades for various types of cancer such as breast cancer [1, 2], colorectal cancer [3, 4], lung cancer [5, 6], and prostate cancer [7]. These signatures use the expression values of several to several hundred selected genes to predict a cancer patient's prognosis. While gene prognostic signatures are crucial in determining personalized treatment, only a few of them have proven values in clinical trials, and are currently used in practice. These include Oncotype DX [8] and MammaPrint [9, 10] for breast cancer. Oncotype DX uses the expression values of 21 genes to calculate a cancer recurrence score (RS), while MammaPrint uses the expression values of 70 genes to classify patients into a high- or low-risk group of recurrence. A critical step of developing these gene signatures was selecting a set of genes that can provide adequate predictive power for cancer prognosis, which remains challenging.

To avoid the step of gene selection, the Cox regression model has been employed to predict prognosis and outcome of cancer based on expression values of a large number of genes [11, 12]. Efficient algorithms have been developed to infer such high-dimensional Cox regression model by maximizing the partial likelihood regularized by the ridge (Cox-ridge), Lasso (Cox-Lasso), or elastic net (Cox-EN) penalty [11]. More recently, artificial neural networks (ANNs) in conjunction with the Cox model have been employed to predict cancer prognosis [13–19]. SurvivalNet [13], DeepSurv [14], and Cox-nnet [15] used a fully connected (FC) ANN or multilayer perceptron (MLP). They provided comparable performance with Cox-ridge or Cox-EN in most types of cancer; SruvivalNet and Cox-nnet offered slightly better performance than Cox-ridge or Cox-EN in one out of three cancers and in two out of ten cancers, respectively.

The performance of ANNs [13–15] was limited possibly by the relatively small number of data samples of each cancer type available to train the model. One approach to mitigating this problem is to employ a method that can exploit data across multiple types of cancer. Towards this end, the meta-learning method [20] and the variational autoencoder (VAE) [21] have been adopted in prediction of cancer prognosis [18, 19]. The meta-learning method achieved comparable or slightly better performance than direct training in lung cancer, glioma, and headneck squamous cell carcinoma [18], while the performance of VAECox in terms of the estimated concordance index (c-index) was slightly better than that of Cox-nnet in 7 out of 10 caners, but it is not clear if difference of the performance reaches any statistical significance.

Contrastive learning (CL) uses a deep ANN to effectively learn feature representations from unlabeled and/or labeled data, which can be used to aid in other learning tasks [22]. In this study, supervised CL [23] is applied to predict cancer prognosis using gene expression values. A classifier is first developed, utilizing features learned by a CL-based ANN to categorize patients into low- or high-risk groups for recurrence. This classification approach is similar to the one used by MammaPrint [9, 10]. A CL-based Cox model (CLCox) is then developed, in which CL is used to learn feature representations that are input to the Cox model for predicting the hazard ratio (HR), which can serve as a prognostic index [24], similar to the RS of Oncotype DX [8], to stratify patients into different risk groups for personalized treatment. Using datasets from The Cancer Genome Atlas (TCGA), it is demonstrated that CL can significantly improve prediction accuracy. Furthermore, CL-based classifiers and CLCox models trained with the TCGA data for lung and prostate cancer are validated by utilizing data from two independent cohorts. Finally, it is shown that the CLCox model trained with expression values of all 13,235 genes in a breast cancer dataset offers significantly better performance than the model trained with expression values of 16 genes of Oncotype DX.

# 2 Results

## 2.1 Overview of machine learning models

As depicted in Fig.1a, our CL-based machine learning models consist of two modules: a CL module to learn feature representations from tumor transcriptomes and a Cox proportional hazards model or a classifier to predict cancer prognosis using the learned features from the CL model. The CL module is a MLP trained with a contrastive loss function [23]. The Cox model was implemented using three existing methods: 1) Cox-EN [11], 2) Cox-XGB that is the gradient boosting-based approach implemented with XGBoost [25], and 3) Cox-nnet that is an ANN-based model [15]. The Cox model was trained using features from the CL module and progress free intervals (PFIs) of the patients. We referred the three CL-based models to as CLCox-EN, CLCox-XGB, and CLCox-nnet, respectively. The Cox model predicts HR that can be used as a prognostic index [24] similar to the RS of Oncotype DX [8]. We also used the classification approach, as used by MammaPrint [9, 10], to classify patients into a high- or low-risk group of cancer recurrence. The gradient boosting classifier implemented with XGBoost [25] was adopted. Patients were divided into two groups: a high risk group with a PFI < 3 years and a low risk group with a PFI ≥ 3 years, and the classifier was trained using features from the CL module and class labels. We referred the CL-based classifier as CL-XGBoost. For performance comparison, we also bypassed the CL module and trained the Cox model and the classifier directly using the gene expression values.

We selected the TCGA datasets of 19 types of cancer, and trained and tested machine learning models for each type of cancer. The names of these 19 cancer types and their abbreviations are given in Methods. Fig. 1b shows the number of patients of each type of cancer. Of note, the number of samples for classification is smaller than that for training the Cox model, because the samples with an censored PFI < 3 years do not have a class label and cannot be used in classification. GBM was not used in classification because the number of samples was too small. For each type of cancer, we used 80% of randomly selected data samples to train the CL model, the Cox model and the classifier, and the remaining 20% data to evaluate the performance of the overall model. Five-fold cross validation (CV) was used to selected the optimal values of hyperparameters of each model and the architecture of the ANN. For performance evaluation, we used Harrell's concordance index (c-index) and the integrated Brier score (IBS) for the Cox model, and the receiver operating characteristic (ROC) curve and the area under the ROC curve (AUC) for the classifier.

While the models trained with the TCGA training data were evaluated with the TCGA test data that were not used in training, we also used data of two independent cohorts to validate the models trained with TCGA data. Specifically, the lung cancer data of the Clinical Proteomic Tumor Analysis Consortium 3 (CPTAC-3) [26] and a prostate cancer dataset named DKFZ [27] were used to validate models trained with TCGA LUAD, LUSC, and PRAD data, respectively. The numbers of samples in the CPTAC-3 and DKFZ datasets are given in Fig.1c.

## 2.2 Contrastive learning improves risk prediction

We trained a classifier to categorize patients into a high- or low-risk group of cancer recurrence, as done by MammaPrint [9, 10]. The architecture of the MLP in the CL module determined by five-fold CV had two hidden layers for all 18 types of cancer. The number of nodes at each hidden layer and the output layer varied for different types of cancer. Specifically, the number of nodes is 5,396 or 5,196 at the first hidden layer, 2,048 or 1,024 at the second hidden layer, and 256 or 128 at the output layer. The feature at the output of the MLP was chosen to train a XGBoost classifier. To see the effect of contrastive learning, we also bypassed the CL module in Fig.1a and trained a XGBoost classifier using the gene expression values as input.

Fig. 2 presents the ROCs and AUCs of the two classifiers with or without CL for 18 types of cancer. It is observed that CL significantly improved the AUC by more than 0.1 for all 18 types of cancer with a p-value less than 0.05 (Wilcoxon rank-sum test). CL increased the AUC by more than 0.2 for 9 types of cancer (CESC, COAD, HNSC, LIHC, OV, SARC, STAD, THCA, and UCEC) and by more than 0.3 for 2 types of cancer (COAD and OV). The AUC of CL-XGBoost exceeds 0.8 for 12 types of cancer, whereas the AUC of XGBoost is less than 0.8 for all 18 types of cancer. Notably, the AUC of CL-XGBoost is 0.906 for LGG and 0.896 for KIRP.

## 2.3 Data pooling improves risk prediction

The clustering analysis of TCGA RNA-seq data by Hoadley *et al.* [28] revealed that gene expression profiles of certain types of cancer were in the same cluster. We reasoned that if the data in the same cluster or group are pooled to train a model, we might improve prediction accuracy especially for those types of cancer with a relatively small number of samples. Towards this end, we identified 8 groups of cancers reported by Hoadley *et al.* [28]. As described in Methods, we used pooled data to train a CL-based XGBoost classifier, which is referred to as CLg-XGBoost, for each of 12 types of cancer belong to at least one group.

Fig. 3 compares the ROCs and AUCs obtained from the pooled data from a group of cancer types and the data of individual type of cancer. Data pooling achieved statistically significant improvement for 9 out of 12 types of cancer except the following three types of cancer: LGG, COAD, and SKCM. In particular, the AUC for BLCA and STAD was improved from 0.790 to 0.885, and from 0.797 to 0.843, respectively. Remarkably, CLg-XGBoost for KIRP and LGG achieved the largest AUC of 0.943 and 0.910, respectively. Results in Figs. 2 and 3 show that CL boosted the AUC of the XGBoost classifier above 0.8 for 14 out of 18 types of cancer.

## 2.4 Contrastive learning improves prediction of the hazard ratio

We compared the performance of six methods that used the Cox model to predict the HR of 19 types of cancer using the TCGA data. These six methods include three existing methods, Cox-nnet, Cox-EN, and Cox-XGB, and the three methods, CLCox-nnet, CLCox-EN, and CLCox-XGB, that we developed to combine CL with the three existing methods. The final MLP in the CL module determined by five-fold CV had two hidden layers. The number of nodes at each layer might be different for different types of cancer. Generally, the first and second hidden layers and the output layer had 5,196 − 5,596, 3,096 − 4,096, and 1,024 − 2,048 nodes, respectively. The features at the output of the first hidden layer was selected to train a Cox model. Five-fold CV was also used to determine the hyper-parameters of the Cox model. In particular, the ANN of Cox-nnet had one hidden layer as also used in [15].

Fig. 4a shows the box plots of c-indexes of the six Cox models trained with TCGA data. It is seen that CL improved the prediction for all 19 types of cancer. To see the performance improvement more clearly, Extended Data Table 1 lists the c-indexes and their standard errors, as well as p-values (Wilcoxon rank-sum test) for the comparison of two methods with and without CL. The increase of the c-index produced by CL is statistically significant (p-value < 0.05) for 55 out of 57 comparisons except the following two comparisons: CLCox-nnet versus Cox-nnet for LIHC and CLCox-XGB versus Cox-XGB for THCA. Averaging across all types of cancer, CL increased the c-index by 0.058, 0.062, and 0.060 for Cox-nnet, Cox-XGB, and Cox-EN, respectively. The largest c-index is 0.880 achieved by CLCox-EN for KIRP. Extended Data Fig. 1 shows the IBS of the six Cox models. It is observed that each CL-based Cox model achieved a smaller or almost the same IBS for every type of cancer comparing with the model without CL.

To see if the HR predicted by a Cox model can distinguish patients with different prognosis, we stratified the patients of a type of cancer into two groups using the median HR in the training set. Fig. 4b depicts the Kaplan-Meier (KM) curves of these two groups of cancer patients in the training and test sets of four types of cancer. The KM curves of the remaining 15 other types of cancer are shown in Extended Data Figs. 2 and 3. The KM curves of the two groups of cancer patients are statistically different (p-value < 0.05, log-rank test) for every type of cancer. This shows the HRs predicted by our models are able to discriminate patients with different prognosis.

## 2.5 Data pooling improves prediction of HR

As described earlier, we considered 8 groups of cancer with similar patterns of gene expression and used the pooled data within a group to train a classifier, which resulted in improved classification accuracy. We also used the pooled data to train Cox models aiming to improve performance. Extended Data Fig. 4 compares the c-indexes of CL models trained with the data of a single cancer or pooled data of multiple types of cancer in a group, and Extended Data Table 2 lists the values of these c-indexes and the p-values (Wilcoxon sum-rank test) for the comparisons. Data pooling provided better or similar c-indexes for almost all 13 types of cancer that belong to at least one group and three Cox models, except for the following five comparisons: CLgCox-nnet vs. CLCox-nnet for HNSC, CLgCox-XGB vs. CLCox-XGB for CESC, HNSC, and LUSC(SM), and CLgCox-EN vs. CLCox-EN for KIRP. Performance improvement is statistically significant (p-value < 0.05) in 20 out of 45 comparisons. In particular, data pooling improved the c-index by more than 0.03 for the following three types of cancer LGG, LUSC, and STAD with at least one Cox model. Moreover, CLgCox models with data pooling achieved a lower or almost the same IBS comparing with CLCox models without data pooling as shown in Extended Data Fig. 4.

## 2.6 Validation on the independent cohorts

We used two independent datasets, the CPTAC-3 lung cancer dataset [26] and the DKFZ prostate cancer dataset [27], to validate classifiers and Cox models trained with the TCGA LUSC, LUAD, and PRAD data. Fig. 5a compares the two ROCs of each classifier for each of the three types of cancer: one obtained with the TCGA test data and the other obtained with CPTAC-3 or DKFZ data. There is no significant difference in 5 out of 6 pairs of AUCs except for that of CL-XGBoost on the PRAD dataset. Although the p-value of CL-XGBoost on the PRAD dataset is 2.90e-06, difference between the two AUCs (0.863 and 0.814) is relatively small. This demonstrates that the performance of the classifiers on independent cohorts is largely not different from that on the original TCGA cohort.

Fig. 5b gives the c-indexes of Cox models obtained with TCGA test data and CPTAC-3 or DKFZ data. Regarding LUSC, c-indexes of each of six Cox models do not exhibit any statistically significant difference for CPTAC-3 and TCGA data. As for LUAD, the CLCox-EN model yielded c-indexes that are not statistically different for CPTAC-3 and TCGA data, while the remaining five models gave statistically different c-indexes on the two datasets. However, the difference of c-indexes between CPTAC-3 and TCGA is relatively small, ranging from 0.017 to 0.043. With respect to the results of PRAD, the p-value for the comparison between the c-indexes of Cox-nnet for TCGA and DKFZ data is 0.019, and that the p-values for the remaining five methods are all greater than 0.05. This indicates that all six Cox models essentially did not produce statistically different c-indexes on TCGA and DKFZ PRAD data. Figure 5c shows the KM curves for the two groups of patients of each of three cancer types, stratified by the median HR predicted by the CLCox-EN model from the TCGA training data. It is observed that the two groups in CPTAC-3 and DKFZ cohorts exhibit statistically different KM curves, similar to that seen in the TCGA test data. Of note, the slight difference in AUCs and c-indexes between the TCGA cohort and the two independent cohorts may be due to the data quality rather than the classifiers, as will be elaborated in Discussion.

## 2.7 The CLCox model with all genes outperforms the model with 16 genes of Oncotype DX

Oncotype DX uses expression values of 16 genes, that are normalized relative to the expression values a set of 5 housekeeping genes, to compute an RS to predict the risk of distant recurrence of ER+ breast cancer patients [8, 29]. Since deep learning can learn features effectively for a specific prediction task [30], we investigated if a CL-based Cox model with expression values of all available genes could offer better performance than the Cox model with the 16 genes of Oncotype DX. The investigation was conducted using a collection of 6 datasets of breast cancer [31, 32] that contain microarray expression data of 13,235 genes and distant metastasis free survival (DMFS) time of 687 estrogen receptor-positive (ER+) patients.

Fig. 6a shows that Cox-EN, Cox-XGB, and Cox-nnet models with the 16 genes of Oncotype DX offer a c-index around 0.65 which is similar to the c-index of the univariate Cox model that uses the RS value of the Oncotype DX as the predictor [32]. The c-index of the Cox-EN model with all 13,235 genes is not statistically different from that of the Cox-EN model with the Oncotype DX genes, whereas the c-indexes of Cox-nnet and Cox-XGB with all genes are slightly higher than those of Cox-nnet and Cox-XGB with Oncotype DX genes with a p-value less than 0.05.

We then trained CLCox-EN, CLCox-XGB, and CLCox-nnet models with 16 genes of Oncotype DX and also with all 13,235 genes. As shown in Fig. 6a, CL improves the c-index for all six models. CLCox-EN, CLCox-XGB, and CLCox-nnet models with all genes offer the best c-index of about 0.72, significantly better than those (about 0.68) of CLCox-EN, CLCox-XGB, and CLCox-nnet models with 16 genes of Oncotype DX, and also those (about 0.65) of the six models without CL. Fig. 6b depicts the KM curves of three risk groups in the test data that were determined using the HRs predicted by the CLCox-EN model in the training data as described in Methods. These KM curves show that three groups of patients have statistically different DMFS probabilities. The CLCox-EN model with all genes exhibits a lower p-value and lower HRs in pair-wise comparisons (high-risk group vs. medium-risk group and medium-risk group vs. low-risk group) than the CLCox-EN model with 16 genes of Oncotype DX, which corroborates its better prediction accuracy as shown in c-indexes in Fig. 6a.

## 3 Discussion

CL is a self-supervised machine learning technique typically carried out with data augmentation that uses the structure of the data such as images to generate similar data instances [22]. Supervised CL has also been developed to take class label information into account, which results in an embedding space where elements of the same class are more closely aligned than in the self-supervised case [23]. Gene expression data do not have the structure of images that can facilitate data augmentation. Therefore, we cannot apply self-supervised CL here, but instead use supervised CL. Since we partitioned data samples into multiple groups with similar PFIs, the low-dimensional features learned by the CL method are closely aligned for those data samples in the same group and more separated for those samples in different groups. The Cox models that we developed to rely on these kind of features were able to offer significantly better performance than existing methods, as observed in our results.

The HR output from the Cox model is a continuous number, similar to the recurrence score of Oncotype DX [8], indicating the risk of cancer recurrence. We also trained a classifier to categorize patients in different risk groups of recurrence, as done by MammaPrint [9, 10]. We demonstrated that CL could significantly improve the accuracy of the classifier. Specifically, for the 18 types of cancer considered, our CL-based classifiers achieved an AUC of between 0.7 and 0.8 for 4 types of cancer, between 0.8 and 0.9 for 12 types of cancer, and above 0.9 for 2 types of cancer, whereas the AUC of the same classifier without using CL was below 0.8 for all 18 types of cancer. The AUC is an indication of the discriminating power of a classifier. The general guidelines of using AUC to determine the discriminating power are as follows [33, 34]: acceptable discrimination ($0.7 \leq$ AUC $<$ 0.8), excellent discrimination ($0.8 \leq$ AUC $< 0.9$ ), and outstanding discrimination (AUC $\geq$ 0.9). Thus, our CL-based classifiers achieved excellent or outstanding AUC for the vast majority of the 18 types of cancer that we analyzed.

We used data of two independent cohorts named CPTAC-3 and DKFZ to validate the Cox models and classifiers trained with the TCGA data of three types of cancer, LUSC, LUAD, and PRAD. Most c-indexes of the Cox models and AUCs of the classifier do not exhibit statistically significant difference on the TCGA data and the CPTAC-3/DKFZ data. For those c-indexes and AUCs that show statistically significant difference, the differences are relatively small. These differences may be due to the data rather than the methods that trained the models. First, we converted the RPKM values in CPTAC-3 and DKFZ RNA-seq data to read counts per gene used in the TCGA data. If the length of a gene used in data conversion is different from the one used in the TCGA data, the converted value may not be consistent with the value in the TCGA data. Second, 534, 561, and 1,215 genes in TCGA LUAD, LUSC, and PRAD data were missing in CPTAC-3 LUAD, CPTAC-3 LUSC, and DKFZ data, respectively. We set the expression values of those missing genes to zero in the validation process. Third, as shown in Fig. 5b, the maximum PFIs in CPTAC-3 and DKFZ datasets are much shorter than those in TCGA data. Despite these factors that might have affected the validation results, the data of independent cohorts largely validate the models trained with TCGA data, demonstrating that our models are relatively robust.

Oncotype DX uses a set of 16 cancer-related genes and 5 reference genes to compute an RS. These 16 genes were carefully selected from 250 candidate genes reported in the literature using the gene expression and clinical data of 447 breast cancer patients [8]. We demonstrated that our CLCox model trained with all 13,235 genes available in a breast cancer dataset provided significantly better performance than the Cox model trained with 16 genes of Oncotype DX. This indicates that the CL-based ANN can learn better features than manually selected 16 genes of Oncotype DX.

The CL-based classifier for each of 18 types of cancer and the CLCox model for each of 19 types of cancer trained with the TCGA RNA-seq data are available as mentioned in the Code Availability section. If the RNA-seq transcriptome of a new tumor of one of these types of cancer is available, one can normalize the RNA-seq data using the 11 housekeeping genes as described in Methods, and then input the normalized data to the corresponding model to predict the recurrence risk of the patient. Python codes for model training are also available, and one can use or modify the codes to train a new CL-based classifier or a CLCox model to possibly improve the prediction of the model, if more data are available. While we used PFI as the clinical endpoint as recommended by Liu *et. al.* [35], one may use a different endpoint such as distant recurrence as used by Oncotype DX [8] and MammaPrint [9, 10] for breast cancer, and retrain classifiers or CLCox models. In short, the trained models and the Python codes in this study provide a valuable resource that will potentially find clinical applications for many types of cancer.

## 4 Methods

### 4.1 Datasets

We used the RNA-seq and clinical data from The Cancer Genome Atlas (TCGA) to train and test models. The datasets are available at the Genomic Data Commons (GDC) of the National Cancer Institute (https://gdc.cancer.gov/about-data/publications/pancanatlas). The RNA-seq dataset contains the expression values of 20,531 genes in 11,069 patients, that have been normalized, batch corrected, platform-corrected by the PanCancer Atlas consortium [28]. Similarly, the corresponding clinical data of the cancer patients have been curated by the PanCancer Atlas consortium [35]. Non-cancer samples were removed, and genes with missing or negative expression values were excluded. Patient samples with missing PFI were also removed. Let the expression value of a gene be denoted as $x$, then, it was log-transformed as $\log_2(1 + x)$. The PFI and the censoring information were extracted from the clinical data. Of note, we used PFI instead of the overall survival (OS) time in our analysis, because PFI is generally a better clinical endpoint choice than OS [35]. We select 18 types of cancer that have > 250 samples with both RNA-seq and PFI data, and glioblastoma multiforme (GBM) that has only 160 patients but a relatively large number of uncensored samples (127), to build and test models for predicting their prognosis. The 18 types of cancer are bladder urothelial carcinoma (BLCA), breast invasive carcinoma (BRCA), cervical squamous cell carcinoma and endocervical adenocarcinoma (CESC), colon adenocarcinoma (COAD), head and neck squamous cell carcinoma (HNSC), kidney renal clear cell carcinoma (KIRC), kidney renal papillary cell carcinoma (KIRP), brain lower grade glioma (LGG), liver hepatocellular carcinoma (LIHC), lung adenocarcinoma (LUAD), lung squamous cell carcinoma (LUSC), ovarian serous cystadenocarcinoma (OV), prostate adenocarcinoma (PRAD), sarcoma (SARC), skin cutaneous melanoma (SKCM), stomach adenocarcinoma (STAD), thyroid carcinoma (THCA), and uterine corpus endometrial carcinoma (UCEC).

We used the data of CPTAC-3 [26] and a prostate cancer data set named DKFZ [27] to validate the models trained with the TCGA data. The CPTAC-3 dataset was downloaded from the GDC data portal. It contains the RNA-seq and clinical data of five types of cancer: LUAD, LUSC, renal cell carcinoma, glioblastoma, and UCEC. Only LUAD and LUSC are among the 18 types of cancer for which a classifier was trained with the TCGA data to predict recurrence risk as described late. Therefore, we extracted RNA-seq data and the PFI information of these two types of cancer, which yielded 206 and 102 samples for LUAD and LUSC, respectively. The DKFZ dataset was downloaded from cBioPortal. It contained RNA-seq and PFI of 105 prostate cancer patients.

A collection of 6 breast cancer microarray datasets was used to train and test Cox models that use the expression values of 16 genes of Oncotype DX [8] or all genes as the input. This collection of data was compiled by van Vliet *et. al.* [31] and was used by Zhao *et. al.* [32] to assess the performance of 9 gene signatures for predicting the prognosis of breast cancer. These 6 datasets were obtained with Affymetrix Human Genome HG-U133A arrays. Five of them can be accessed from the Gene Expression Omnibus (GEO) with the following accession numbers: GSE6532 [36], GSE3494 [37], GSE1456 [38], GSE7390 [39], and GSE2603 [40]; the other one can be accessed from ArrayExpress with the accession number E-TABM-158 [41]. We obtained from the author of [32] the preprocessed and normalized gene expression data and clinical data that include DMFS time. The pooled dataset contains a total of 947 patients, and we extracted 687 ER+ samples. It has the expression values of 20,3507 probes; we mapped these probes to 13,235 genes and calculated the expression value of each gene as the mean of expression values of all probes mapped to the gene.

## 4.2 CL-based classifier

Our CL-based classifier consists of two modules: a CL module that uses an ANN to learn a representation of the high dimensional gene expression vector of a patient in a low dimensional space, and a classification module that uses the features from the CL module to classify patients into a low- or high-risk of recurrence. The CL module is an MLP with sigmoid activation at hidden layers, and we used the supervised contrastive loss function [23] in training. Suppose that the training data set contains $n$ patients of a particular type of cancer. Let the vector $\mathbf{x}_i$ represent the expression values of genes in patient $i$, and $t_i$ be the PFI of patient $i$. Without loss of generality, let us assume $t_1 \leq t_2 \leq \cdots \leq t_n$. We divided $n$ samples into $m$ groups, each of which contains $\lfloor n/m \rfloor$ or $\lfloor n/m \rfloor + 1$ samples, where $\lfloor n/m \rfloor$ stands for the largest integer that is less than or equal to $n/m$. We chose $m$ such that $\lfloor n/m \rfloor$ was around 15. For each $\mathbf{x}_i$, we obtained a class label $y_i$ to be the index of the group that it belongs to.

During the training, we took a mini-batch of $n_b$ samples denoted as $\{\tilde{\mathbf{x}}_i, \tilde{y}_i, i = 1, 2, \cdots, n_b\}$. Let us define the set $I := \{1, 2, \cdots, n_b\}$. If we randomly select an $i \in I$, let $A(i):=I\setminus\{i\}$ be the set that contains the elements of $I$ excluding $i$. For a sample $(\tilde{\mathbf{x}}_i, \tilde{y}_i)$ in the mini-batch, we define a positive set that contains all the samples with the same label as $\tilde{\mathbf{x}}_i$: $P(i) := \{p \in A(i) : \tilde{y}_p = \tilde{y}_i\}$. Let $\mathbf{z}_i = MLP(\mathbf{x}_i)$ be the output the neural network given the input $\mathbf{x}_i$. Then, we used the following contrastive loss function in training the neural network [23]:

$$\mathcal{L} = \sum_{i \in I} \mathcal{L}_i = \sum_{i \in I} \frac{-1}{|P(i)|} \sum_{p \in P(i)} \log \frac{\exp(\mathbf{z}_i^T \mathbf{z}_p / \tau)}{\sum_{a \in A(i)} \exp(\mathbf{z}_i^T \mathbf{z}_a / \tau)},$$

where $|P(i)|$ is the cardinality of the set $P(i)$, and $\tau$ is a positive constant.

We randomly split the dataset of each type of cancer into a training set (80%) and a test set (20%). The training set was used to train the CL module. More specifically, the training data were used to train MLPs with the contrastive loss function. MLPs were built in PyTorch and trained using the SGD algorithm with $\ell_2$-regularization and early stopping. The hyperparameters include the following: the number of layers (2,3,4,5), the number of nodes at each layer (1,024 - 8,192), the learning rate ($10^{-4}, 5 \times 10^{-5}, 2 \times 10^{-5}, 10^{-5}, 10^{-6}$), and the weight for the $\ell_2$-regularization term (0.01, 0.005, 0.003, 0.001, 0.0007, 0.0003, and 0.0001). Five-fold CV was employed to search over the space of the hyperparameters, and five MLPs with the smallest validation loss were selected. Training samples ($\mathbf{x}_i$'s) were input to each of MLP, and the output of each MLP at a hidden layer or the output layer ($\mathbf{z}_i$'s) were used to train a classifier to categorize patients into a low- or high-risk group of recurrence.

To train the classifier, we divided all patients into two groups: a high-risk group with PFI < 3 years and a low-risk group with PFI > 3 years, and labeled the available data samples correspondingly. Of note, those samples that were censored and whose PFI was < 3 years could not be labeled, and therefore they were removed from the data. For all the samples in the training set, we got their outputs from the CL model and used them to train a gradient boosting classifier, which was implemented with XGBoost [25]. Five-fold CV was employed to choose one of the five MLPs and one of the hidden layers or the output layer in the CL module and the optimal hyperparameters of the XGBoost classifier including the size of each tree, the number of trees, and the learning rate. The test data were applied to the trained CL model and the classifier, and their class labels were predicted. The predicted results were used to plot a ROC curve and to calculate AUC as the performance measure. The above procedure of randomly splitting data, training and testing the model was repeated 40 times, and the mean and standard deviation of AUCs were calculated, and the Wilcoxon rank-sum test was conducted for performance comparison.

## 4.3 The CLCox method

Similar to the CL-based classifier, our CLCox method consists of two modules: a CL module for learning feature representations and a Cox module that uses the features from the CL model to predict the prognosis of cancer. We used the Cox proportional hazards model in the Cox module. To predict caner prognosis, we used PFI as the clinical endpoint as recommended by Liu *et. al.* [35]. Let us define the hazard function $h(t)$ of a cancer patient as the instantaneous potential per unit time for the event of cancer recurrence to occur at a time $t$ given that the patient is free of cancer up to $t$. The Cox proportional hazards assumes $h(t)$ in a semi-parametric form [42]: $h(t) = h_0(t)e^{f_\theta(\mathbf{x})}$, where $\mathbf{x}$ is a vector consisting of all covariates or features which are gene expression values in our case, $h_0(t)$ is a baseline hazard function that does not depend on $\mathbf{x}$, $f_\theta(\mathbf{x})$ is the output of the model, and $\theta$ represents all the model parameters. We used three existing methods for the Cox model: 1) the regularized Cox regression model with the elastic net (EN) penalty [11], 2) the gradient boosting-based approach implemented with XGBoost [25], and 3) the neural network-based model, Cox-nnet [15]. We refer to the first and the second methods as Cox-EN and Cox-XGB, respectively. The output $f_\theta(\mathbf{x})$ was modeled with a linear function of $\mathbf{x}$ in Cox-EN, a sum of numerous trees in Cox-XGB, and a neural network in Cox-nnet. The parameters of the Cox model can be estimated by maximizing the partial likelihood, which is given by $L(\theta) = \prod_{i=1}^n \left( \frac{\exp(f_\theta(\mathbf{x}))}{\sum_{j \in R(t_i)} \exp(f_\theta(\mathbf{x}))} \right)^{\delta_i}$, where $n$ is the number of individuals, $R(t_i)$ is the set of individuals that are at risk of experiencing an event at time $t_i$, and $\delta_i$ indicates whether individual $i$ is terminated by an event ($\delta_i = 1$) or by censoring ($\delta_i = 0$).

To train and test CLCox for each selected type of cancer, we randomly divided the data into a train set with 80% samples and a test set with 20% samples. The CL module was trained with the same procedure as the one described earlier for the CL-based classifier. To train the Cox model, the negative log partial likelihood was used as the loss function. Five-fold CV was used to search over the space of the hyperparameters of the Cox model, five MLPs, and the hidden layer of each MLP whose output was used as the features for training the Cox model. More detailed description of training three Cox models is described in the next section. Of note, the test data were never used in training the CL module and the Cox model.

## 4.4 Implementation of Cox-EN, Cox-XGB, and Cox-nnet

Cox-EN was implemented with Python module scikit-survival, and Cox-XGB was implemented with the software package XGBoost [25]. The regularization term of Cox-EN is $\lambda(\alpha\|\boldsymbol{\theta}\|_1 + (1-\alpha)/2\|\boldsymbol{\theta}\|_2^2)$, where $\|\boldsymbol{\theta}\|_1$ and $\|\boldsymbol{\theta}\|_2$ are the $\ell_1$- and $\ell_2$-norm of $\boldsymbol{\theta}$, respectively, and $\lambda > 0$ and $0 \leq \alpha \leq 1$ are two hyperparameters. Five-fold cross validation was employed to search over the two pre-specified sets of values for $\lambda$ and $\alpha$ to determine their optimal values. The XGBoost model has a number of hyperparameters. We used the default values of most hyperparameters except for the maximum tree depth, parameter $\lambda$ for $\ell_2$-regularization, and subsample, which were determined with five-fold CV.

We downloaded the Python code of Cox-nnet [15] and updated several Python functions so that the code could run under Python 3.11 and PyTorch 2.0. We trained the ANNs of Cox-nnet, that have one or two fully connected (FC) hidden layers with sigmoid action function, with $\ell_2$-regularization by minimizing the Cox loss function, which is the negative log partial likelihood. The hyperparameters include the number of hidden layers (1, 2), the number of nodes at hidden layer(s), the weight of $\ell_2$-regularization, and the learning rate. Five-fold CV was employed to search over the set of hyper-parameter values to determine the optimal neural network.

## 4.5 Evaluation of Cox models

The performance of Cox models on prognosis prediction was evaluated with two commonly used metrics: the Harrell's concordance index (C-index) [43] and the integrated Brier score (IBS) [44]. The C-index is defined as the ratio of the concordant pairs of predictions and all comparable pairs of patients, where a pair of patients is called concordant if the risk of the event predicted by a model is lower for the patient who experiences the event at a later time point. A C-index value of 0.5 indicates random prediction. The value of C-index increases when the prediction accuracy increases, and a value of 1.0 indicates perfect prediction, where all pairs are concordant.

Suppose that the test set contains $m$ samples and let $t_i$ and $\delta_i$, $i = 1, 2, \cdots, m$ be the PFI and the indicator of the censoring status, respectively, of the $i$th sample. The Brier score at time $t$ under random censorship can be estimated as [44, 45]

$$BS(t) = \frac{1}{m} \sum_{i=1}^{m} \left\{ \frac{(0 - \hat{S}_i(t))^2 1_{t_i \leq t, \delta_i = 1}}{\hat{G}(t_i)} + \frac{(1 - \hat{S}_i(t))^2 1_{t_i > t}}{\hat{G}(t)} \right\},$$

where $S_i(t)$ is the estimated survival function of individual $i$, and $\hat{G}(t)$ is the Kaplan-Meier estimated of the censoring distribution. The survival function $S_i(t)$ can be estimated as $\hat{S}_i(t) = e^{-\hat{H}_0(t) \exp(f_\theta(\mathbf{x}_i))}$, where $\hat{H}_0(t)$ is the cumulative baseline hazard function estimated from the training data using the Breslow estimator [46]:

$$\hat{H}_0(t) = \sum_{j: t_j \leq t} \left\{ \frac{\delta_j}{\sum_{k \in \mathcal{R}(t_j)} \exp(f_\theta(\mathbf{x}_k))} \right\}.$$

The IBS is then calculated as IBS $= \frac{1}{T} \int_0^T BS(t) dt$, where $T$ is determined as follows. It is seen from the formula of $\hat{H}_0(t)$ that we can get $\hat{H}_0(t)$ up to the maximum uncensored $t_j$ in the training data. However, the size of the set $\mathcal{R}(t_j)$ decreases when $t_j$ increases. For those $t$ where the formula of $\hat{H}_0(t)$ contains $\mathcal{R}(t_j)$ whose size is small, $\hat{H}_0(t)$ may not be accurate. To avoid this problem, we define $t_{\max}$ as the maximum uncensored $t_j$ in the training data where the size of $\mathcal{R}(t_j)$ is greater than or equal to 20. Also, we define $\tilde{t}_{max}$ as the maximum $t_j$ in the test data. Then, we set $T = \min\{t_{\max}, \tilde{t}_{\max}\}$.

As mentioned earlier, we randomly split data into a training set and a test set. We used the training set to train the model, used the trained model to make prediction on the test set, and then computed the two performance metrics. We repeated this process 40 times, each time with a random seed for splitting the data. We used results of 40 repeats to produce box plots of the two performance metrics, to compute their mean and standard deviation, and to perform Wilcoxon rank-sum test for performance comparison.

## 4.6 Pooling data of different types of cancer for model training

We identified the following 8 groups of cancers from clustering analysis of TCGA RNA-seq data by Hoadley *et al.* [28]: a group with squamous morphology (BLCA, CESC, ESCA, HNSC, and LUSC), glioma tumors (GBM and LGG), melanomas of the skin and eye (SKCM and UVM), clear cell and papillary renal carcinomas (KIRC and KIRP), hepatocellular and cholangiocarcinomas (LIHC and CHOL), a gastrointestinal group (COAD, READ, non-squamous ESCA, READ, and STAD), a digestive system group (PAAD, STAD, and a few ESCA), and two mixed lung cancer groups (LUAD and LUSC). For each group of cancers, we split the data into a training set with 80% samples and a test set with 20% samples, and used the training data of all types of cancers in the group to train the MLP in the CL module. Then, for each type of cancer in the group, we used the features output from the MLP with the samples in the training set of that type of cancer to train a XGBoost classifier or a Cox model. We again considered three Cox models: Cox-EN, Cox-XGB, and Cox-nnet. Of note, the test data were not used in training the CL model, the classifier, or the Cox model.

## 4.7 Model validation with independent cohorts

We used CPTAC-3 lung cancer (LUAD and LUSC) data [26] and DKFZ prostate cancer data [27] to validate the Cox models and classifiers trained with TCGA LUAD, LUSC, and PRAD data. As mentioned earlier, the CPTAC-3 dataset contains 206 LUAD patients and 102 LUSC patients, and the DKFZ dataset contains 105 PRAD patients, which are used to validate Cox models trained with TCGA data. To validate the classifiers, we divided data samples for each type of cancer into a high-risk group (PFI < 3 years) and a low-risk group (PFI > 3 years). This resulted in 47, 125, 23 samples in the high-risk group, and 45, 53, 47 samples in the low-risk group for CPTAC-3 LUSC, CPTAC-3 LUAD, and DKFZ PRAD datasets, respectively. Of note, the samples with a censored PFI < 3 years had to be discarded, because their class labels were unknown.

Both CPTAC-3 and DKFZ RNA-seq data contain reads per kilobase million (RPKM) values for each gene. But the pan-cancer TCGA RNA-seq data that we used contain the read counts for each gene (per private communication with the first author of [28]). Therefore, we converted RPKM values to read counts per gene by multiplying the RPKM value of a gene by the length of the gene.

We used expression values of housekeeping genes to normalize gene expression values in different datasets. Based on RNA-seq data, Eisenberg and Levanon identified eleven genes that were highly uniform and strongly expressed in all human tissues [47]. Ten out of the eleven genes are present in the TCGA dataset, and they are C1orf43, CHMP2A, GPI, PSMB2, PSMB4, RAB7A, REEP5, SNRPD3, VCP, and VPS29. We computed the average expression values of these 10 genes in the TCGA and CPTAC-3 LUAD datasets, which are denoted as $E_t$ and $E_c$, respectively. Then, the expression values of all genes in the CPTAC-3 LUAD dataset were multiplied by a normalization factor $E_t/E_c$. Similarly, we used the 10 housekeeping genes to normalize the gene expression values of the LUSC datasets. We found that the variances of the expression values of the 10 housekeeping genes in the TCGA PRAD dataset had relatively large variations. Therefore, we selected the three genes, VCP, RAB7A, and GPI, that had smallest variance in their expression values, and used the average expression values of these three genes in the TCGA PRAD and DKFZ datasets to normalize gene expression values.

Not all genes in the TCGA dataset are present in the CPTAC-3 and DKFZ datasets. Specifically, 534, 561, and 1,215 genes in TCGA LUAD, LUSC, and PRAD data were missing in CPTAC-3 LUAD, CPTAC-3 LUSC, and DKFZ data, respectively. Of note, the number of genes in TCGA LUAD and LUSC data is 20,531, but the number of genes in TCGA PRAD data is 16,136, because some genes with missing expression values were removed. Therefore, we extracted expression values of 19,997, 19,970, and 14,921 genes in CPTAC-2 LUAD and LUSC, and DKFZ datasets, respectively, that are present in the corresponding TCGA datasets. The expression values of these genes were $\log_2$-transformed as done to the TCGA gene expression values, and expression values of the missing genes were set to zero. Gene expression values of the CPTAC-3 and DKFZ data were input to the Cox models and classifiers trained with the corresponding TCGA LUAD, LUSC, and PRAD data, and the outputs of each model were used to compute performance metrics such as c-index and IBS for Cox models and AUC for classifiers.

### 4.8 Comparison of Cox models with Oncotype DX genes and all available genes

We used gene expression and DMFS data of 687 ER+ breast patients in a collection of 6 breast cancer datasets [31, 32] described earlier to train Cox models. Expression values of each gene were normalized by subtracting the average value of the following five housekeeping genes: ACTB, GAPDH, RPLPO, GUS, and TFRC [8]. The dataset was randomly split into a training set with 80% samples and a test set with 20% samples. We trained Cox-EN, Cox-XGB, Cox-nnet, CLCox-EN, CLCox-XGB, and CLCox-nnet models with the 16 genes of Oncotype DX, and also trained these six models with all 13,235 genes, and then determined the c-index of each model using the test data. This process of random splitting of data, training, and testing was repeated 40 times.

Oncotype DX uses the RS to divide patients into three groups: a high-risk group (RS≥ 31), a medium-risk group (18≤ RS< 31), and a low-risk group (RS<18) [8]. Clinical trials showed that patients in different groups had different DMFS probabilities [8, 29]. We used the HR predicted by a Cox model to divide patients into three groups. Specifically, we ranked HRs of all patients, predicted by Cox-EN with the training data, in the descent order, and followed the approach in [8, 32] to determine two cutoff values $c_1$ (= 73th percentile of HRs) and $c_2$ (= 51st percentile of HRs). We divided the patients in the test data into three groups: a high-risk group (HR≥ $c_1$), a medium-risk group ($c_2$ ≤ HR< $c_1$), and a low-risk group (HR< $c_1$), and then compared the KM curves of the patients in the three groups.

## 5 Data availability

All data used in this paper are publicly available. The TCGA pan-cancer RNA-seq and clinical data were downloaded from the GDC website: https://gdc.cancer.gov/about-data/publications/pancanatlas. The CPTAC-3 dataset was obtained from the GDC data portal at https://portal.gdc.cancer.gov/projects/CPTAC-3. The DKFZ prostate cancer dataset was downloaded from the cBioPortal website at https://www.cbioportal.org/. Five microarray datasets can be accessed from the GEO database at http://www.ncbi.nlm.nih.gov/geo using the GEO accession numbers mentioned in the Methods section, and another microarray dataset can be accessed from ArrayExpress at http://www.ebi.ac.uk/arrayexpress using the accession number mentioned in the Methods section.

# 6 Code availability

Python codes for training and testing all the models in this paper are publicly available at the following GitHub site: https://github.com/CaixdLab/CL4CaPro. CL-based classifiers and CLCox models trained with the TCGA data are freely accessible at the following Box link: https://miami.box.com/s/ylmvqynbtchx5xhof0quaeu9w62mxaca. We have documented software libraries and provided sample codes for replicating the results in the paper, training and testing new models, and using the trained models to make predictions.

47. Eisenberg, E. & Levanon, E. Y. Human housekeeping genes, revisited. *TRENDS in Genetics* **29,** 569–574 (2013).

## Acknowledgments

Research reported in this publication was supported by the Sylvester Comprehensive Cancer Center in partnership with the University of Miami College of Engineering under the Engineering Cancer Cures program Award Number SCCC-ECC-2022-01.

## Author contributions

X.C. conceived the methods and designs of the study. Z.C. contributed to designs. A.S. made computer programs and conducted the computational experiments. X.C., A.S., E.J.F., and Z.C. analyzed the results. X.C. drafted the manuscript. All authors reviewed the manuscript and approved the final version.

## Competing interests

The authors declare no competing interests.

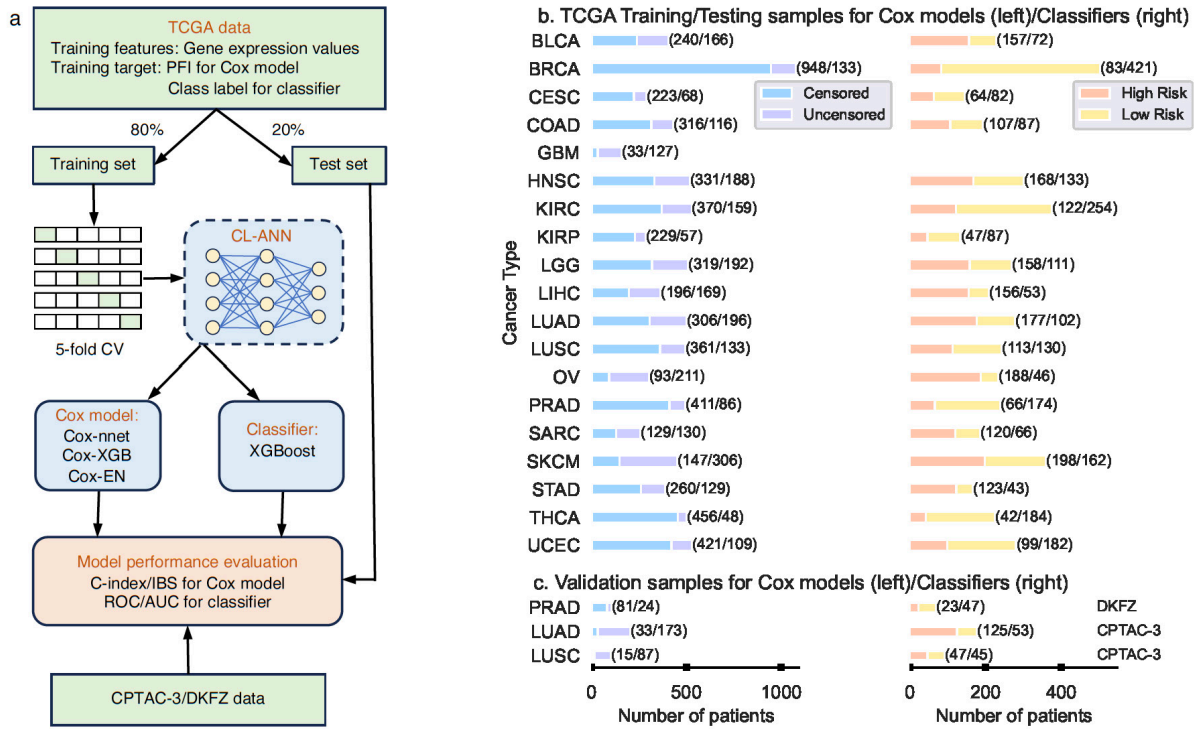

Figure 1: **Overview of development of machine learning models for predicting cancer prognosis.** **a.** Flow chart for training and testing Cox models and classifiers. For each of 19 types of cancer, TCGA gene expression and clinical data were randomly split into a training set (80% samples) and a test set (20% samples). The training data were used to train a contrastive learning-based artificial neural network (CL-ANN). The features learned by CL-ANN were used to train a Cox model to predict hazard ratio or a classifier to categorize patients into a high or low risk group of disease recurrence. Five-fold cross validation was employed to determine optimal hyperparameters of the model. Test data were then used to evaluate the performance of the trained model. Data from CPTAC-3 and DKFZ were further used to validate the models of LUAD, LUSC, and PRAD trained with TCGA data. This process of random splitting of data, training, testing, and validating each model was repeated 40 times. **b.** List of 19 types of cancer and the numbers of TCGA samples for training and testing Cox models (left) and classifiers (right). **c.** Numbers of samples of lung cancer (LUAD and LUSC) and prostate cancer (PRAD) from two independent cohorts, CPTAC-3 and DKFZ, for validating the Cox models (left) and classifiers (right) trained with TCGA data.

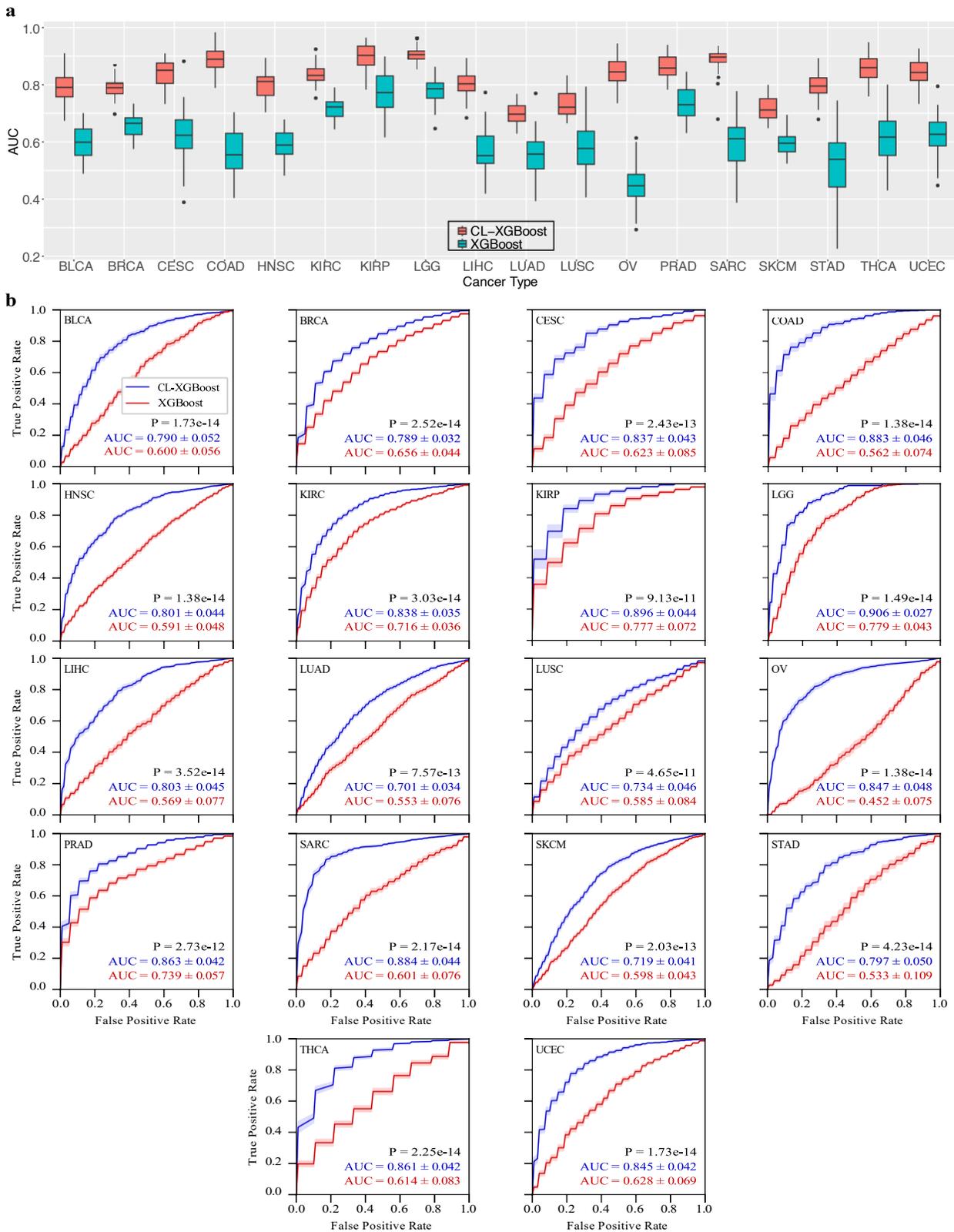

Figure 2: **Performance of the XGBoost classifier with and without contrastive learning for classifying cancer patients into a high or low risk group of disease recurrence. a.** Box plots of AUCs. CL-XGBoost represents the XGBoost classifier that uses the features learned by the contrastive learning module. **b.** ROCs of CL-XGBoost and XGBoost for 18 types of cancer.

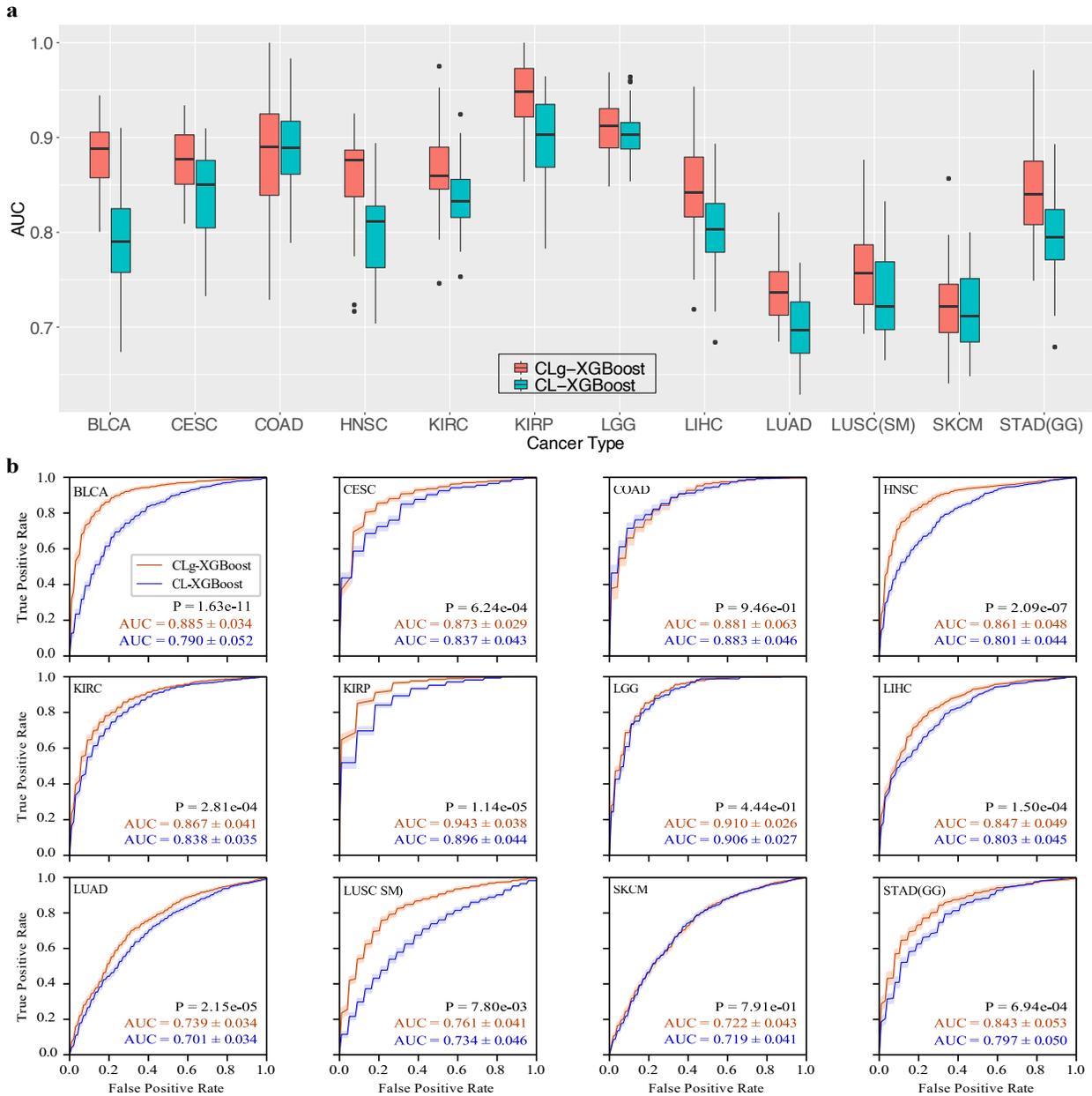

Figure 3: **Performance of contrastive learning classifiers trained with the data of a single type of cancer and the pooled data of a group of different types of cancer.** **a.** Box plots of AUCs for 12 types of cancer, each of which belongs to at least one of the 9 groups. CLg-XGBoost represents the CL-XGBoost classifier trained with the pooled data. LUSC(SM) and STAD(GG) standard for the LUSC and STAD classifiers trained with CL features learned from the data in the squamous morphology (SM) group and the gastrointestinal group (GG), respectively. **b.** ROCs of CLg-XGBoost and CL-XGBoost for 12 types of cancer.

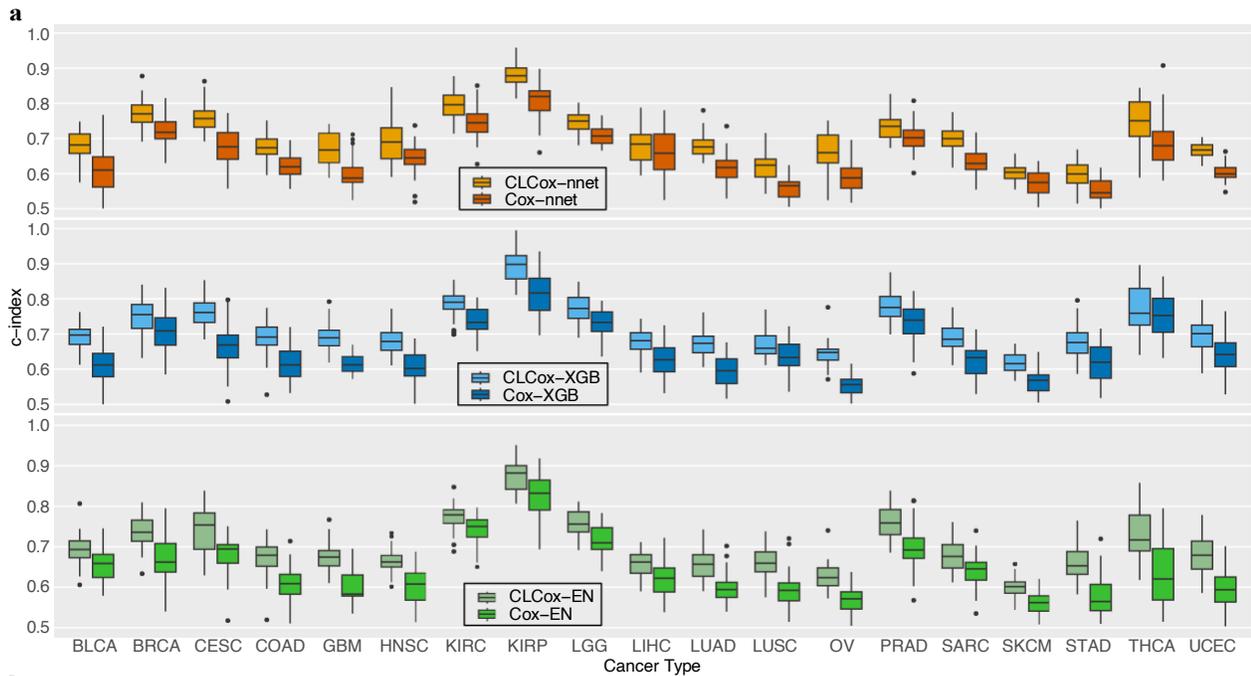

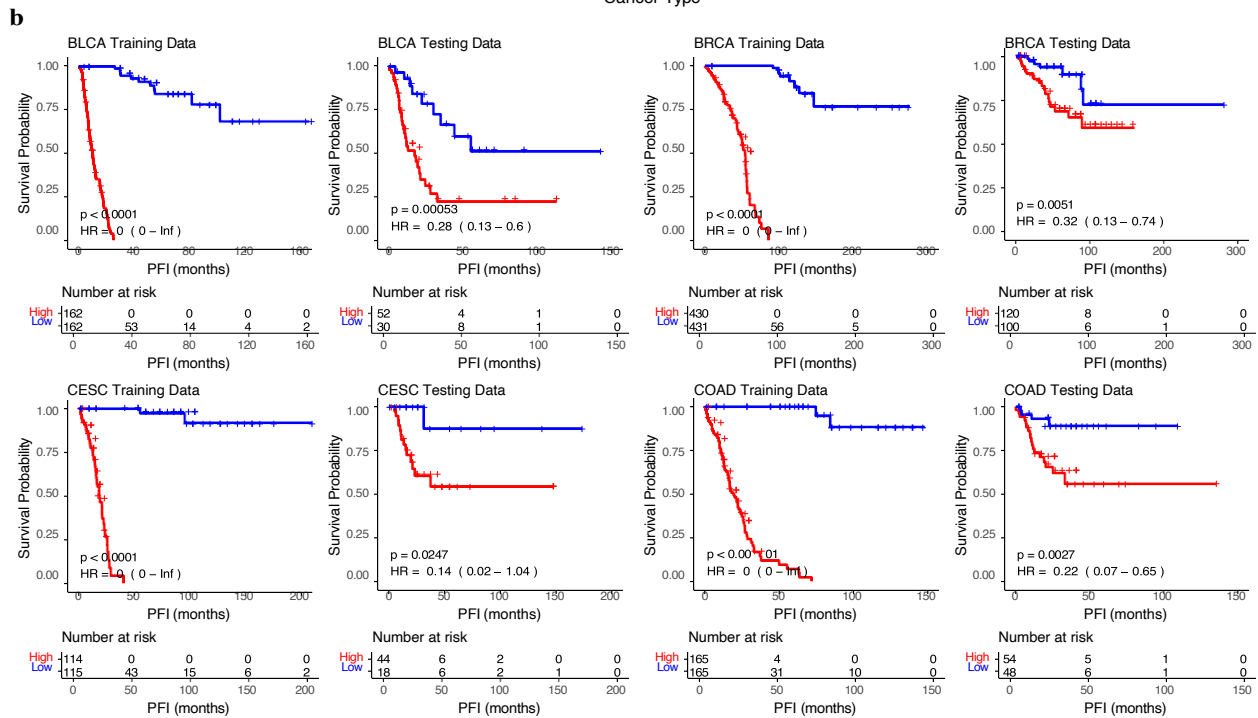

Figure 4: **Performance of Cox models in predicting the hazard ratio of 19 types of cancer using the TCGA data. a.** Box-plots of c-indexes of Cox models. **b.** KM-curves for the two groups of patients stratified by the median HR predicted by the CLCox-XGB model: a poorly prognostic group (red) with patients' HR greater than the median HR and a better prognostic group (blue) with patients' HR less than the median HR. The KM-curves for the remaining 15 types of cancer are presented in Extended Data Figs. 2 and 3.

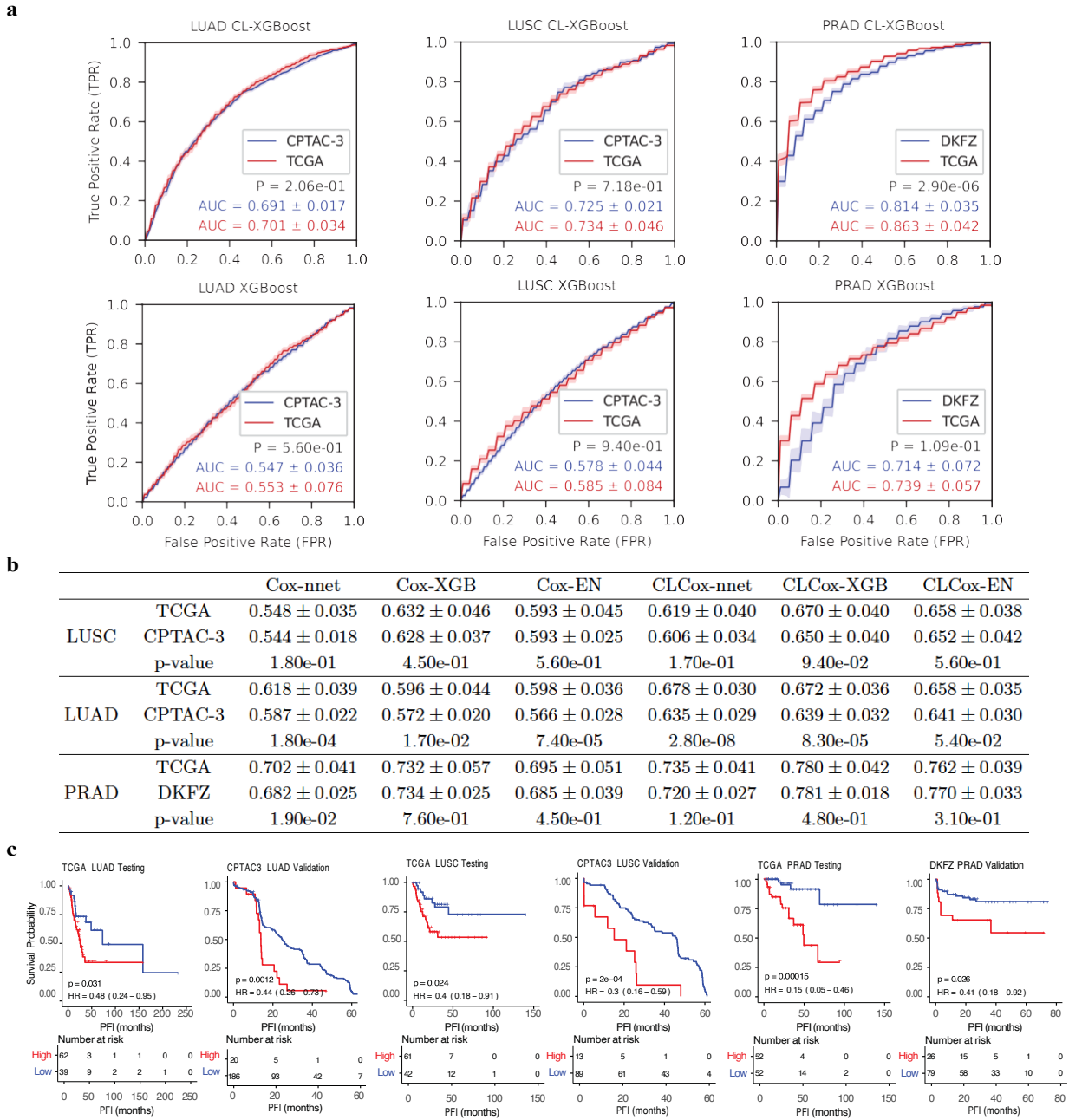

Figure 5: **Validation of Cox models and classifiers trained with the TCGA data using two independent cohorts (CPTAC-3 and DKFZ).** **a.** ROC curves of CL-XGBoost and XGBoost. **b.** C-indexes of six Cox models. **c.** KM curves for the two groups stratified by the median HR predicted by the CLCox-EN model from the TCGA training data: a high-risk group (red) with HRs > the median HR and a low-risk group (blue) with HRs < the median HR.

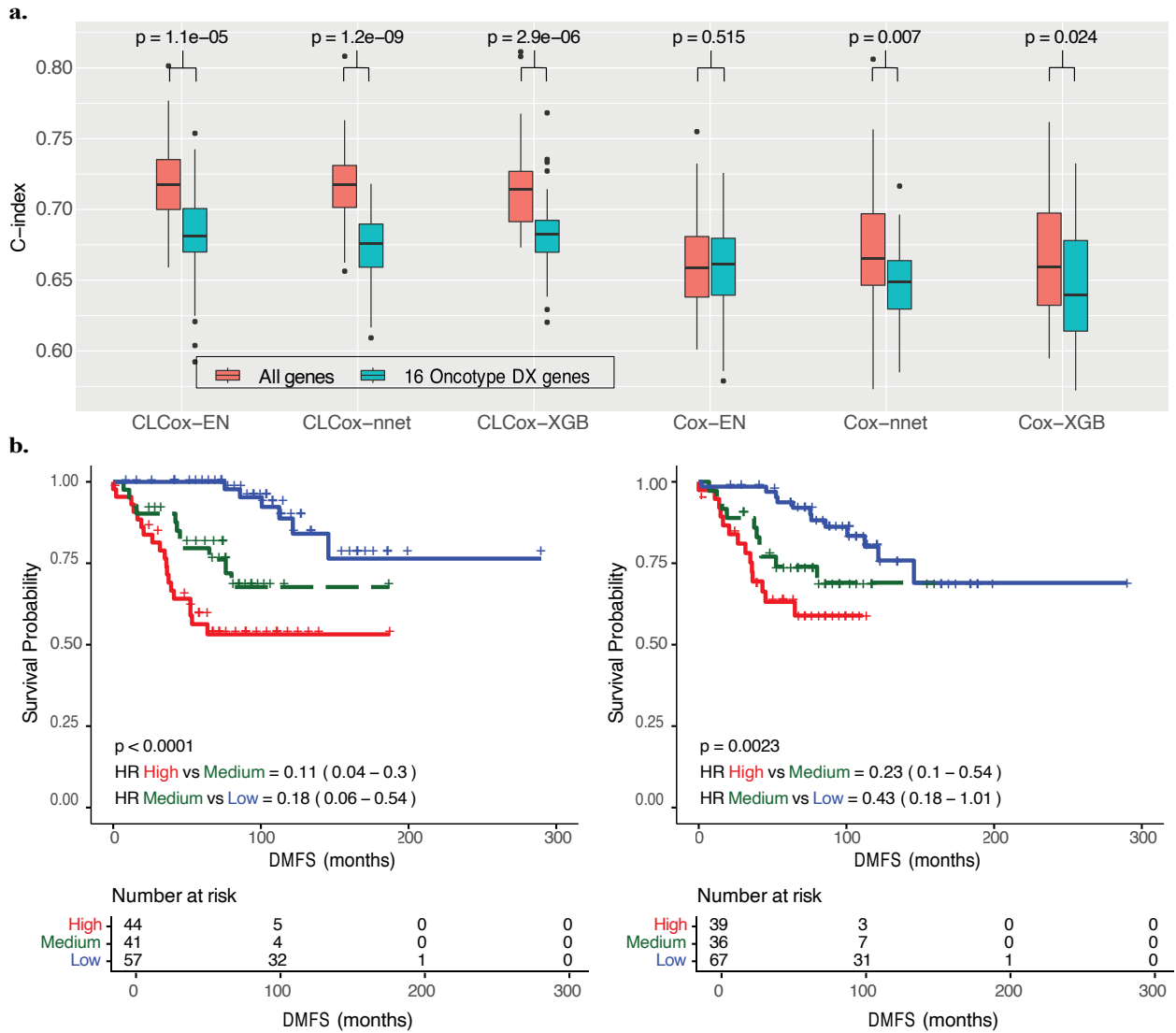

Figure 6: **Performance of Cox models in predicting the distant recurrence risk of breast cancer. a.** Box-plots of c-indexes of Cox models. **b.** KM-curves for the three groups of patients stratified by the HR predicted by the CLCox-EN model with all 13,235 genes (left) and with 16 genes of Oncotype DX (right): a high-risk group (red) with patients' HR greater than the 73th percentile of all HRs, a low-risk group prognostic group (blue) with patients' HR less than the 51st percentile of all HRs, and a medium-risk group with patients' HR in between the 51st percentile and the 73th percentile.

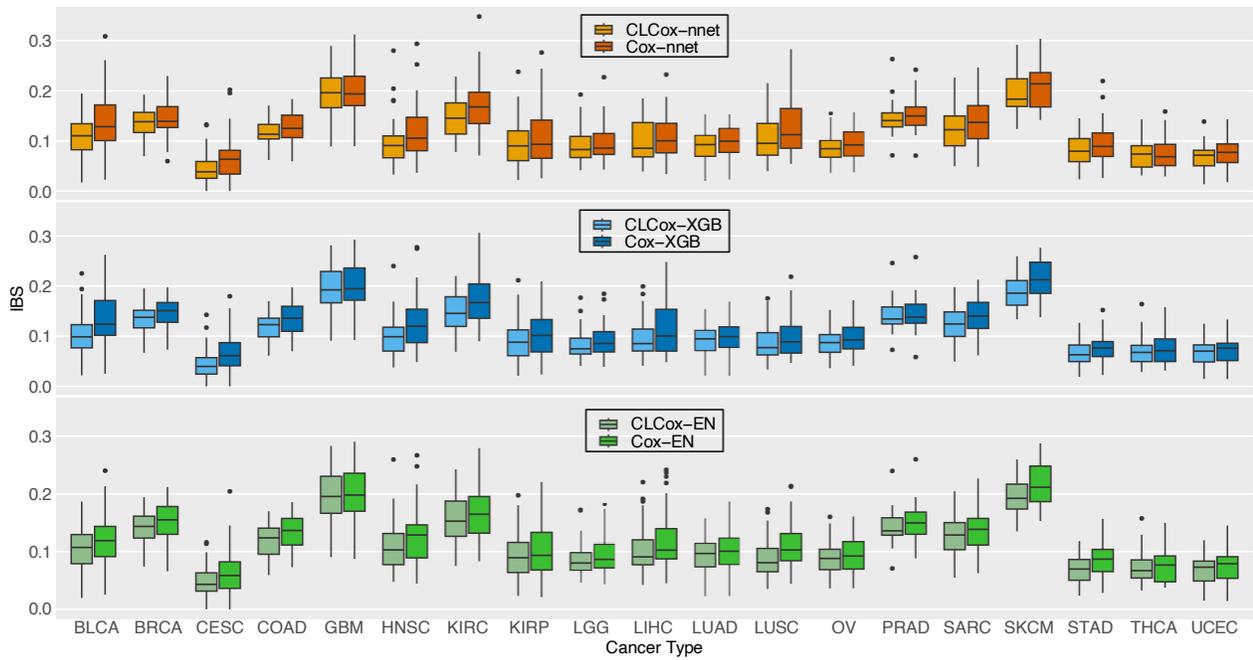

**Extended Data Figure 1**: IBSs of Cox models, each of which was trained with the data of one of 19 types of cancer.

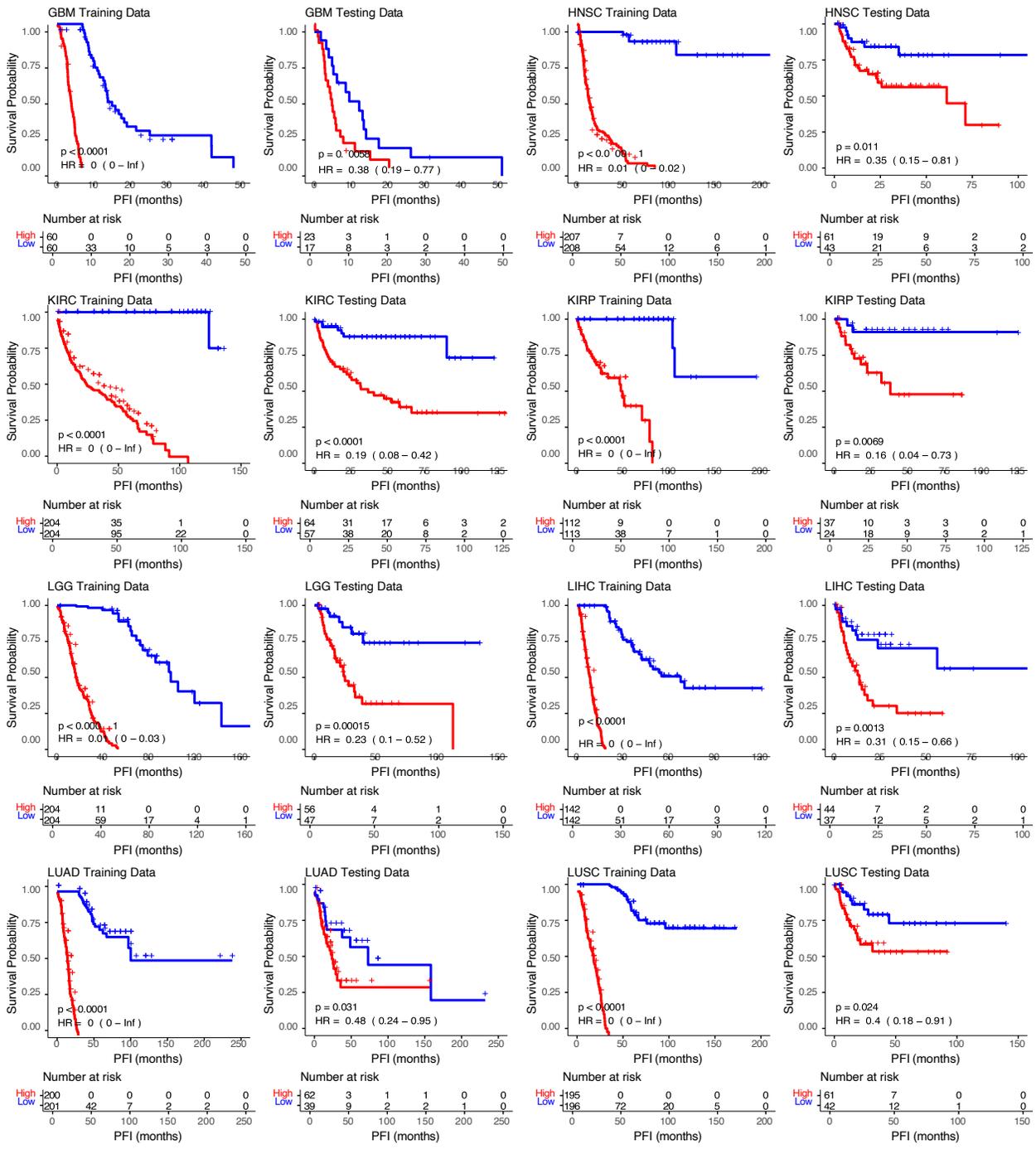

**Extended Data Figure 2**: **KM-curves for the two groups of patients stratified by the median HR predicted by the CLCox-XGB model.** The two groups are a poorly prognostic group (red) with patients' HR greater than the median HR and a better prognostic group (blue) with patients' HR less than the median HR.

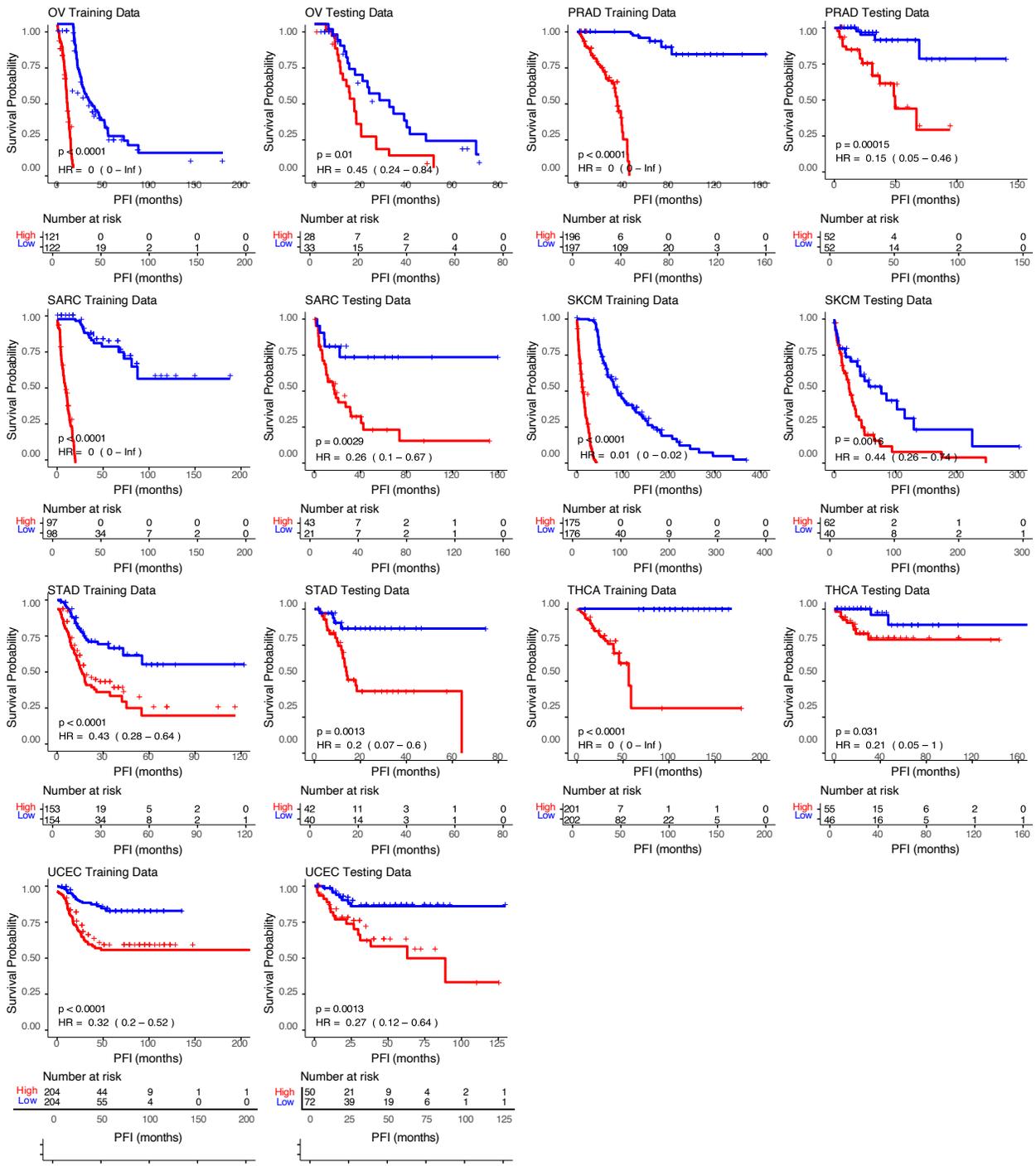

**Extended Data Figure 3**: **KM-curves for the two groups of patients stratified by the median HR predicted by the CLCox-XGB model.** The two groups are a poorly prognostic group (red) with patients' HR greater than the median HR and a better prognostic group (blue) with patients' HR less than the median HR.

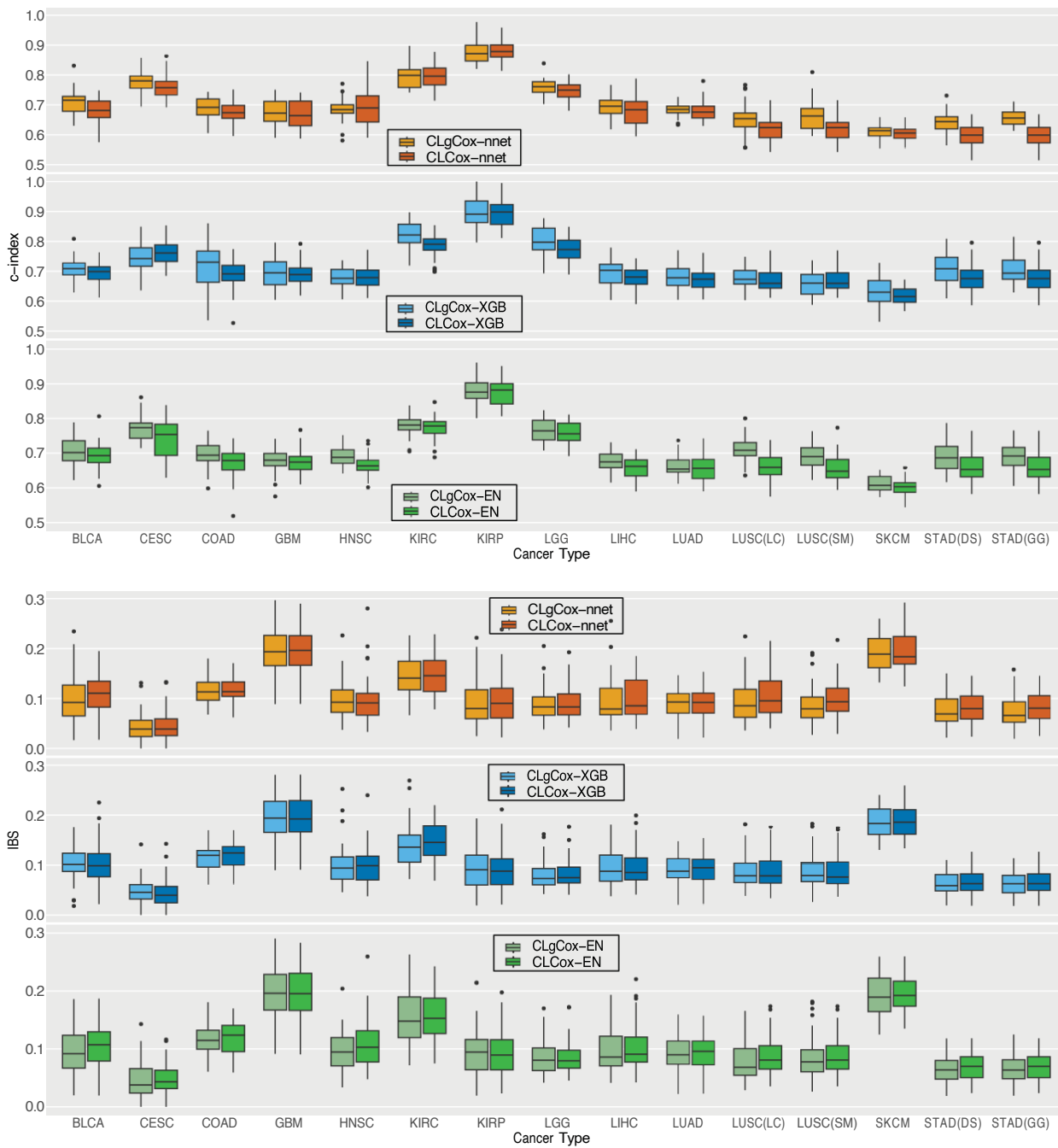

**Extended Data Figure 4**: **Performance of Cox models trained with pooled data of a group of cancer types and data of a single type of cancer.** CLgCox-EN, CLgCox-XGB, and CLgCox-nnet stands for the methods trained with pooled data. LUSC(SM) and LUSC(LC) represent LUSC trained within the group with squamous morphology and the mixed lung cancer group, respectively. STAD(GG) and STAD(DS) represent STAD trained within the gastrointestinal group and the digestive system group, respectively.

Extended Data Table 1: C-indexes of six Cox models in predicting the hazard ratio of 19 types of cancer.

| Cancer Type | CLCox-mnet | Cox-mnet | p-value | CLCox-XGB | Cox-XGB | p-value | CLCox-EN | Cox-EN | p-value |
|---|---|---|---|---|---|---|---|---|---|
| BLCA | 0.681 ± 0.040 | 0.603 ± 0.067 | 2.4e-07 | 0.693 ± 0.037 | 0.610 ± 0.051 | 2.7e-10 | 0.692 ± 0.035 | 0.653 ± 0.041 | 8.3e-05 |
| BRCA | 0.772 ± 0.044 | 0.720 ± 0.044 | 5.6e-06 | 0.753 ± 0.047 | 0.708 ± 0.053 | 2.2e-04 | 0.737 ± 0.040 | 0.672 ± 0.058 | 7.6e-07 |
| CESC | 0.756 ± 0.037 | 0.679 ± 0.050 | 9.4e-10 | 0.761 ± 0.040 | 0.664 ± 0.062 | 1.1e-10 | 0.742 ± 0.058 | 0.679 ± 0.055 | 2.3e-05 |
| COAD | 0.677 ± 0.033 | 0.619 ± 0.042 | 1.4e-08 | 0.688 ± 0.048 | 0.613 ± 0.052 | 1.7e-08 | 0.671 ± 0.041 | 0.605 ± 0.044 | 4.1e-09 |
| GBM | 0.670 ± 0.048 | 0.600 ± 0.041 | 7.8e-09 | 0.690 ± 0.037 | 0.616 ± 0.028 | 4.9e-12 | 0.677 ± 0.035 | 0.607 ± 0.046 | 3.1e-08 |
| HNSC | 0.696 ± 0.058 | 0.643 ± 0.043 | 1.4e-04 | 0.682 ± 0.038 | 0.607 ± 0.043 | 5.4e-10 | 0.663 ± 0.028 | 0.602 ± 0.047 | 2.0e-08 |
| KIRC | 0.795 ± 0.038 | 0.751 ± 0.049 | 3.4e-05 | 0.787 ± 0.040 | 0.734 ± 0.047 | 2.3e-06 | 0.772 ± 0.031 | 0.742 ± 0.033 | 8.0e-05 |
| KIRP | 0.879 ± 0.033 | 0.811 ± 0.049 | 2.3e-09 | 0.894 ± 0.048 | 0.817 ± 0.064 | 5.6e-07 | 0.880 ± 0.039 | 0.827 ± 0.054 | 1.9e-05 |
| LGG | 0.746 ± 0.027 | 0.708 ± 0.027 | 1.3e-07 | 0.773 ± 0.039 | 0.733 ± 0.035 | 4.0e-05 | 0.758 ± 0.032 | 0.717 ± 0.038 | 1.0e-05 |
| LIHC | 0.680 ± 0.049 | 0.655 ± 0.070 | 1.3e-01 | 0.678 ± 0.034 | 0.625 ± 0.051 | 3.1e-06 | 0.658 ± 0.031 | 0.619 ± 0.038 | 8.0e-06 |
| LUAD | 0.678 ± 0.030 | 0.617 ± 0.039 | 1.8e-10 | 0.672 ± 0.036 | 0.595 ± 0.044 | 2.7e-10 | 0.657 ± 0.035 | 0.598 ± 0.037 | 2.6e-09 |
| LUSC | 0.619 ± 0.040 | 0.549 ± 0.035 | 1.4e-10 | 0.670 ± 0.040 | 0.633 ± 0.046 | 1.0e-03 | 0.661 ± 0.040 | 0.593 ± 0.046 | 1.4e-08 |
| OV | 0.664 ± 0.051 | 0.584 ± 0.052 | 9.8e-09 | 0.643 ± 0.035 | 0.550 ± 0.030 | 6.6e-14 | 0.627 ± 0.032 | 0.565 ± 0.038 | 2.1e-10 |
| PRAD | 0.735 ± 0.041 | 0.703 ± 0.041 | 9.3e-04 | 0.780 ± 0.043 | 0.732 ± 0.057 | 3.4e-04 | 0.762 ± 0.039 | 0.695 ± 0.051 | 2.0e-08 |
| SARC | 0.698 ± 0.034 | 0.634 ± 0.041 | 5.5e-09 | 0.691 ± 0.039 | 0.623 ± 0.044 | 3.9e-09 | 0.678 ± 0.038 | 0.638 ± 0.043 | 1.8e-04 |
| SKCM | 0.601 ± 0.024 | 0.567 ± 0.037 | 2.2e-05 | 0.617 ± 0.028 | 0.559 ± 0.039 | 1.9e-09 | 0.602 ± 0.027 | 0.560 ± 0.026 | 7.8e-09 |
| STAD | 0.599 ± 0.034 | 0.536 ± 0.044 | 1.4e-08 | 0.676 ± 0.047 | 0.618 ± 0.054 | 1.0e-05 | 0.659 ± 0.044 | 0.565 ± 0.056 | 4.0e-10 |
| THCA | 0.746 ± 0.067 | 0.684 ± 0.068 | 1.1e-04 | 0.771 ± 0.065 | 0.752 ± 0.062 | 2.5e-01 | 0.727 ± 0.066 | 0.631 ± 0.084 | 3.1e-06 |
| UCEC | 0.667 ± 0.020 | 0.604 ± 0.025 | 3.5e-13 | 0.694 ± 0.046 | 0.644 ± 0.054 | 4.9e-05 | 0.678 ± 0.044 | 0.588 ± 0.055 | 6.1e-10 |

Extended Data Table 2: Comparison of c-indexs of Cox models trained with and without data pooling.

| Cancer Type | CLgCox-nnet | CLCox-nnet | p-value | CLgCox-XGB | CLCox-XGB | p-value | CLgCox-EN | CLCox-EN | p-value |
|---|---|---|---|---|---|---|---|---|---|
| BLCA | 0.710 ± 0.040 | 0.681 ± 0.040 | 7.1e-03 | 0.704 ± 0.038 | 0.693 ± 0.037 | 1.9e-01 | 0.708 ± 0.043 | 0.692 ± 0.035 | 1.2e-01 |
| CESC | 0.776 ± 0.036 | 0.756 ± 0.037 | 9.9e-03 | 0.748 ± 0.053 | 0.761 ± 0.040 | 2.1e-01 | 0.770 ± 0.034 | 0.742 ± 0.058 | 7.8e-02 |
| COAD | 0.692 ± 0.033 | 0.677 ± 0.033 | 3.6e-02 | 0.719 ± 0.071 | 0.688 ± 0.048 | 1.4e-02 | 0.697 ± 0.038 | 0.671 ± 0.041 | 7.9e-03 |
| GBM | 0.675 ± 0.042 | 0.670 ± 0.048 | 6.4e-01 | 0.696 ± 0.049 | 0.690 ± 0.037 | 4.9e-01 | 0.680 ± 0.034 | 0.677 ± 0.035 | 2.8e-01 |
| HNSC | 0.687 ± 0.036 | 0.696 ± 0.058 | 6.2e-01 | 0.680 ± 0.031 | 0.682 ± 0.038 | 9.8e-01 | 0.689 ± 0.026 | 0.663 ± 0.028 | 5.5e-05 |
| KIRC | 0.793 ± 0.035 | 0.795 ± 0.038 | 7.4e-01 | 0.819 ± 0.047 | 0.787 ± 0.040 | 1.8e-03 | 0.780 ± 0.031 | 0.772 ± 0.031 | 2.9e-01 |
| KIRP | 0.879 ± 0.041 | 0.879 ± 0.033 | 5.9e-01 | 0.896 ± 0.054 | 0.894 ± 0.048 | 8.7e-01 | 0.879 ± 0.041 | 0.880 ± 0.039 | 8.0e-01 |
| LGG | 0.759 ± 0.027 | 0.746 ± 0.027 | 5.1e-02 | 0.805 ± 0.046 | 0.773 ± 0.039 | 1.7e-03 | 0.766 ± 0.033 | 0.758 ± 0.032 | 3.8e-01 |
| LIHC | 0.694 ± 0.033 | 0.680 ± 0.049 | 1.4e-01 | 0.695 ± 0.042 | 0.678 ± 0.034 | 3.5e-02 | 0.675 ± 0.027 | 0.658 ± 0.031 | 1.8e-02 |
| LUAD | 0.683 ± 0.021 | 0.678 ± 0.030 | 1.5e-01 | 0.685 ± 0.042 | 0.672 ± 0.036 | 1.9e-01 | 0.661 ± 0.030 | 0.657 ± 0.035 | 6.8e-01 |
| LUSC(LC) | 0.654 ± 0.052 | 0.619 ± 0.040 | 1.0e-03 | 0.676 ± 0.033 | 0.670 ± 0.040 | 2.6e-01 | 0.710 ± 0.034 | 0.661 ± 0.040 | 6.2e-07 |
| LUSC(SM) | 0.660 ± 0.047 | 0.619 ± 0.040 | 2.7e-04 | 0.660 ± 0.042 | 0.670 ± 0.040 | 3.3e-01 | 0.692 ± 0.034 | 0.661 ± 0.040 | 6.8e-04 |
| SKCM | 0.610 ± 0.023 | 0.601 ± 0.024 | 1.1e-01 | 0.634 ± 0.044 | 0.617 ± 0.028 | 8.3e-02 | 0.612 ± 0.023 | 0.602 ± 0.027 | 1.5e-01 |
| STAD(DS) | 0.641 ± 0.038 | 0.599 ± 0.034 | 5.1e-06 | 0.708 ± 0.056 | 0.676 ± 0.047 | 1.3e-02 | 0.689 ± 0.042 | 0.659 ± 0.044 | 2.5e-03 |
| STAD(GG) | 0.657 ± 0.028 | 0.599 ± 0.034 | 1.2e-10 | 0.709 ± 0.048 | 0.676 ± 0.047 | 3.3e-03 | 0.690 ± 0.040 | 0.659 ± 0.044 | 1.5e-03 |